\pdfoutput=1
\documentclass{article}

\usepackage[preprint]{neurips_2024}

\usepackage[utf8]{inputenc} 
\usepackage[T1]{fontenc}    
\usepackage{hyperref}       
\usepackage{url}            
\usepackage{booktabs}       
\usepackage{amsfonts}       
\usepackage{nicefrac}       
\usepackage{microtype}      
\usepackage[table,xcdraw]{xcolor}         

\usepackage{tabularx,booktabs}
\newcolumntype{Y}{>{\centering\arraybackslash}X}

\usepackage{bm}
\usepackage{balance}
\usepackage{graphicx}
\usepackage{multirow}
\usepackage{xspace}
\usepackage{threeparttable}
\usepackage{enumitem}
\usepackage{makecell}
\usepackage{float}
\usepackage{wrapfig}
\usepackage[export]{adjustbox}

\usepackage{amsmath,amsfonts,bm}

\def\eqref#1{equation~\ref{#1}}

\def\1{\bm{1}}

\def\vh{{\bm{h}}}

\def\vw{{\bm{w}}}
\def\vx{{\bm{x}}}
\def\vy{{\bm{y}}}

\def\mA{{\bm{A}}}

\def\mW{{\bm{W}}}
\def\mX{{\bm{X}}}

\DeclareMathAlphabet{\mathsfit}{\encodingdefault}{\sfdefault}{m}{sl}
\SetMathAlphabet{\mathsfit}{bold}{\encodingdefault}{\sfdefault}{bx}{n}

\newcommand{\vpara}[1]{\vspace{0.04in}\noindent\textbf{#1}\xspace}

\newcommand{\model}{GraphAlign\xspace}
\newcommand{\hide}[1]{}

\title{\model: Pretraining One Graph Neural Network on Multiple Graphs via Feature Alignment}

\author{Zhenyu Hou$^{*}$, Haozhan Li$^{*}$, Yukuo Cen, Jie Tang, Yuxiao Dong \\\{houzy21, hz-li20\}@mails.tsinghua.edu.cn}

\begin{document}

\maketitle

\renewcommand{\thefootnote}{\fnsymbol{footnote}}
    \footnotetext[1]{ZH and HL contribute equally.}
\renewcommand{\thefootnote}{\arabic{footnote}}

\begin{abstract}
Graph self-supervised learning (SSL) holds considerable promise for mining and learning with graph-structured data. 
Yet, a significant challenge in graph SSL lies in the feature discrepancy among graphs across different domains. 
In this work, we aim to pretrain one graph neural network (GNN) on a varied collection of graphs endowed with rich node features and subsequently apply the pretrained GNN to unseen graphs. 
We present a general \model method that can be seamlessly integrated into the existing graph SSL framework.  
To align feature distributions across disparate graphs, \model designs alignment strategies of feature encoding, normalization, alongside a mixture-of-feature-expert module. 
Extensive experiments show that \model empowers existing graph SSL frameworks to pretrain a unified and powerful GNN across multiple graphs, showcasing performance superiority on both in-domain and out-of-domain graphs. 
\end{abstract}

\section{Introduction}

\begin{figure}[ht]
    \centering
    \begin{minipage}[t]{0.47\linewidth}
        \includegraphics[width=\linewidth]{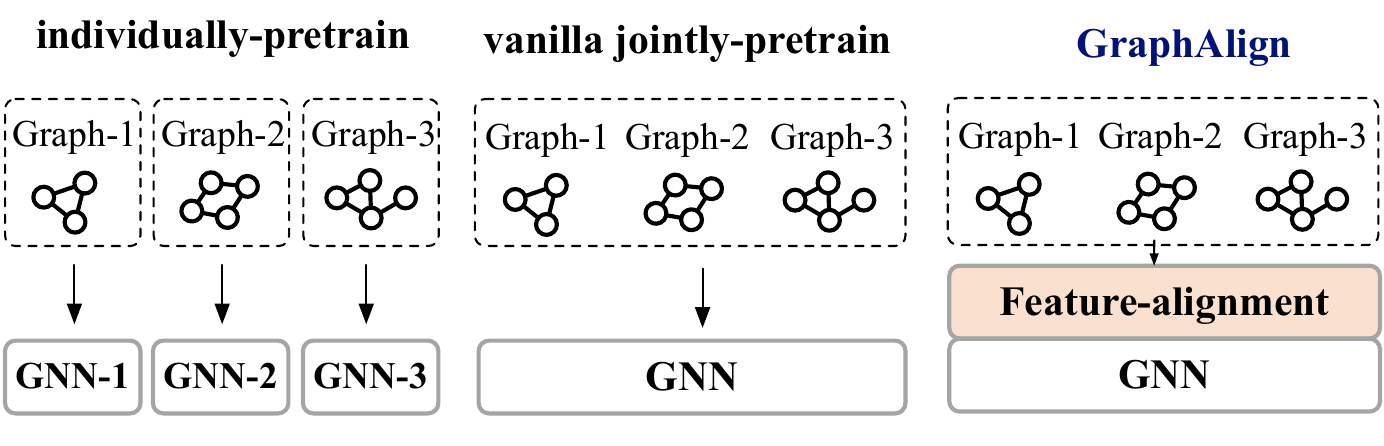}
    \end{minipage}
    \begin{minipage}[t]{0.47\linewidth}
        \includegraphics[width=\linewidth]{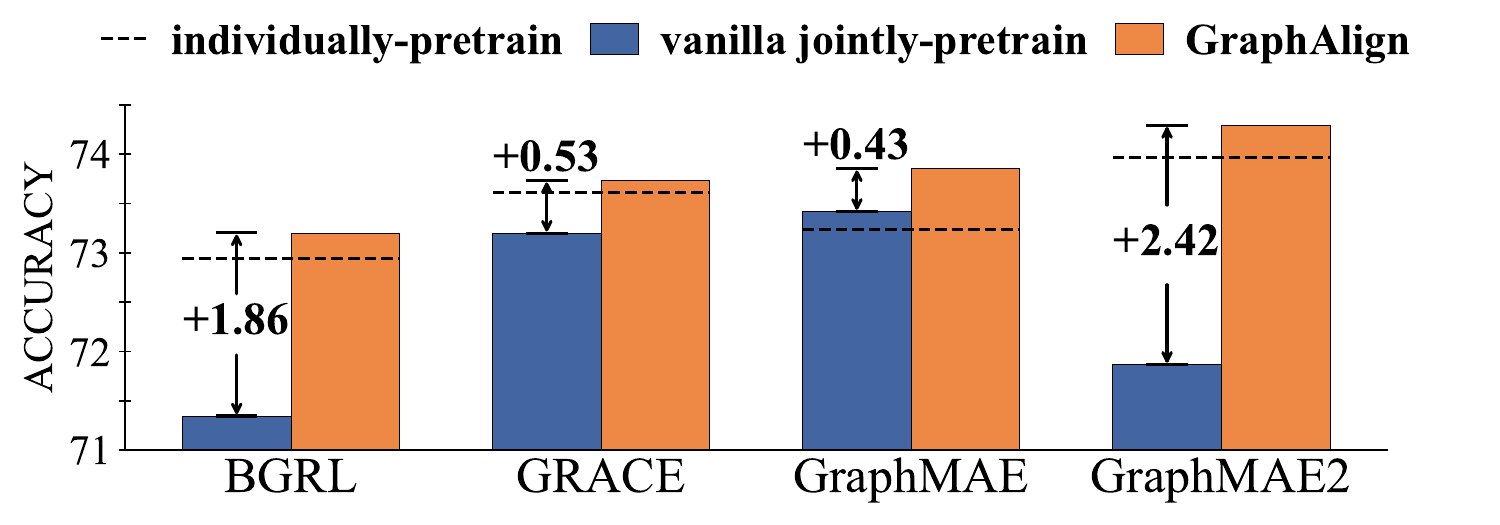}
    \end{minipage}
    \caption{Linear probing results on OGB node classification datasets. \textmd{\textit{individually-pretrain} denotes that we train an individual GNN for each dataset. \textit{vanilla jointly-pretrain} represents training one GNN using all datasets without the incorporation of any designs. \model empowers us to train one GNN that can achieve superior performance across various datasets and shows a clear advantage over vanilla jointly-pretraining.}}
    \label{fig:polit_exp}
\end{figure}

Graph neural networks (GNNs)~\cite{gcn,GAT,gin} have emerged as a cornerstone in modeling graph-structured data, finding applications across various domains ranging from social networks~\cite{gnn4social_network} to recommender systems~\cite{gnn4recommend_survey}. 
The advent of self-supervised learning (SSL) in GNNs has opened avenues for leveraging unlabeled data to capture latent features. 
As the development and apparent potential of pretrained models, developing a universal and powerful graph pretrained model has raised significant attention and 
exploration.

However, the pursuit for a unified and universally-transferable pretrained GNN has encountered various challenges, primarily due to the inherent heterogeneity in graph data. 
Two of the most prominent aspects focus on structure-transfer and feature-transfer across graphs. 
Structure-transfer aims to discover the structural patterns exhibited by graphs from different domains, such as cycles, sparsity, or connectivity patterns. 
Previous works~\cite{gcc,zhu2021transfer,rolx_gccrelated} have posited the potential universality and transferability of structural patterns and achieved promising results on graph structure prediction.

Feature-transfer for graph pretraining on the other hand has thus far remained largely unexplored, given the challenges stemming from the disparate nature of feature representations across graph datasets. 
Specifically, graphs from different sources often encapsulate features that reside in vastly different semantic spaces. 
This discrepancy is not merely a matter of domain variation, such as academic graphs versus product graphs, but extends to the semantic interpretation of the features themselves, like numerical values and continuous embeddings from word2vec~\cite{word2vec}. 
Such diversity in feature representation exacerbates the difficulty in establishing a unified training procedure.

Recently, there are attempts~\cite{liu2023one,huang2023prodigy,allinone} to design elaborated graph prompts for specific tasks such as graph classification. 
This approach usually utilizes additional task-specific nodes as prompts to blend subgraph-level features, thereby fostering the transferability of GNNs. 
Despite its feasibility in supervised-learning~\cite{supervised_learning} or meta-learning~\cite{meta_learn} settings, it cannot address the critical issue of feature misalignment between disparate graphs, thus hindering its applications in the (task-agnostic) self-supervised pretraining context. 
Thus, there is still a long-standing gap between graph pretraining and the powerful pretrained models in natural language and vision domains~\cite{bert,gpt3,he2022masked}. 


\vpara{Contributions.} In this paper,  
we aim to pretrain one GNN on top of a (diverse) set of graphs with rich node features in a self-supervised manner. 
The pretrained GNN can be then applied to downstream graphs and tasks unseen during training. 
The idea is to align feature distributions across different graphs. 
We propose a general \model method that can be straightforwardly used in existing graph SSL frameworks such as BGRL~\cite{BGRL}, GRACE~\cite{grace}, and GraphMAE~\cite{graphmae}. 
As shown in Figure~\ref{fig:polit_exp}, existing SSL frameworks with \model are enabled with a unified pretraining across multiple graphs to achieve performance improvements, while jointly training a single GNN without feature alignment undermines SSL performance. 

Central to \model are three coupled components: feature encoder, feature normalization, and mixture-of-feature-expert projector. 
First, we leverage a language model as the feature encoder to translate the textual attributes of nodes into a shared semantic space. 
It ensures that textual features are sufficiently generalized. 

Second, given that graphs from various domains tend to form distinct clusters in the representation space, we employ feature normalization to reduce semantic disparities. 
This technique standardizes the feature distributions of all graphs to a mean of zero through a centering process applied individually to each graph. 
It also reduces the difficulty for GNNs to model diverse distributions. 

Third, to further capture nuances in node- and graph-specific features, we design a \textit{mixture-of-feature-expert} module---inspired by the mixture-of-experts (MoE)~\cite{shazeer2017outrageously} model---that is positioned prior to the GNN layers. 
This module consists of a routing gate and $K$ different feature projectors.
The routing gate can dynamically assign nodes to distinct feature transformation experts based on their characteristics, i.e., node features rather than its source graph.
The feature experts are then expected to map nuanced distributions into a common input distribution for subsequent GNNs. 
Consequently, this design enables GNNs to process a unified distribution, thereby enhancing their capacities to discern inter-graph commonalities and facilitate transferability.

Additionally, we present a simple and effective few-shot strategy that does not require newly elaborated modules. 
This differs from previous works that usually employ either an extra GNN~\cite{huang2023prodigy} or graph prompting designs~\cite{liu2023one} to enable in-context inference with GNNs. 
Our in-context inference strategy is task-agnostic and can be combined with existing graph self-supervised methods.

We conduct extensive experiments to demonstrate the performance and transferability of the proposed \model. 
The results indicate that \model can achieve better performance than individual training or naive joint training in both generative and contrastive SSL tasks. 
And it also shows promising transferability and achieves competitive or state-of-the-art results in representation learning and few-shot classification benchmarks.

\section{Related Work}

\subsection{Graph self-supervised learning}
Graph self-supervised learning (SSL) aims to pretrain a graph neural network without task-specific supervision. Generally, graph SSL yields a pretrained GNN or generates embeddings for all nodes.
It encompasses both contrastive and generative approaches.
Contrastive methods~\cite{velivckovic2018deep,sun2019infograph,BGRL,hassani2020contrastive,zhang2021canonical} focus on maximizing mutual information and leveraging graph views for representation learning, empirically heavily depending on elaborated graph augmentation operations. 
Generative methods~\cite{kipf2016variational,pan2018adversarially,wang2017mgae,tang2022graph} generally focus on reconstructing graph structures and node attributes. And masked graph autoencoders~\cite{graphmae,graphmae2,hu2020gpt} have shown more promising performance in various graphs tasks.

However, all of them conduct pretraining on a single graph and make no attempts to link different graphs or cross-domain tasks. This is one main difference between graph pretraining and large language model training, which motivates us to explore the power of the unified pretrained GNN.

\subsection{Training unified graph models} 
In the past year, there have been plenty of efforts~\cite{chen2024exploring,huang2023prodigy,liu2023one,allinone,he2023harnessing}, aiming to leverage one trained GNN to handle graphs from different domains and multiple tasks. Almost all of them focus on text-attributed graphs to avoid the discrepancy in the semantic space of node features.
Prodigy~\cite{huang2023prodigy} focuses on enabling in-context learning for GNNs by applying a prompt graph representation that connects examples and queries, allowing models to perform new classification tasks on unseen graphs without fine-tuning. It pays attention to in-domain transfer yet doesn't consider cross-domain tasks. 
One-for-all~\cite{liu2023one}
uses text-attributed graphs to integrate diverse graph datasets and employs a graph prompting paradigm for in-context learning across classification tasks.
All-in-one~\cite{allinone} proposes
reformulates tasks to a graph-level format and utilizes meta-learning to improve multi-task performance. 
Both One-for-all and All-in-one require complex prompt designs during training and inference to align different graphs and achieve transferability.

Yet one main limitation is that there works have to rely on abundant task-specific labels for training. In this paper, we focus on data perspective and architecture design to tackle the challenges. And especially, our method is under a self-supervised setting and is compatible with all existing graph self-supervised learning methods, which is more applicable in real scenarios and could be more promising in graph pretraining.

\begin{figure}[ht]
\centering
\includegraphics[width=\textwidth]{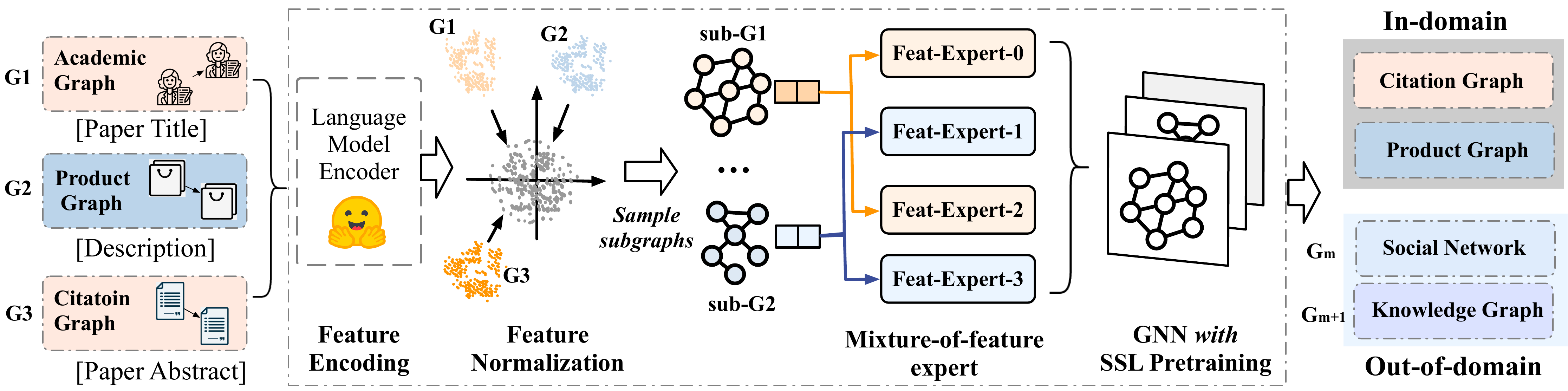}
\caption{Overview of \model. \textmd{Given graphs from different domains, we first utilize a language model functioning as an encoder to project their node attributes into a semantic dimension. 
Subsequently, we apply feature normalization to each graph individually. 
To further capture subtle differences, a mixture-of-feature-experts module is designed and implemented to allow each node to adaptively select feature transformations. 
Finally, we are enabled to pretrain a unified GNN on this aligned feature distribution with any self-supervised learning methods. 
This GNN can be applied to downstream graphs and tasks unseen during training.}}  
\label{fig:overview}
\end{figure}

\section{Method}
\label{section:Method}
In this section, we present the \model framework with the goal of pretraining one unified graph neural network across graphs from devious domains. 
The idea is to design feature alignment strategies to effectively integrate different graphs and bring mutual benefits in the self-supervised setting. 

In this work, we focus on text-attributed graphs (TAGs). 
TAGs are defined as a type of graph in which each node is linked to textual features. 
Let $\mathcal{G} = (\mathcal{V}, \mX, \mA)$ denote a graph, where $\mathcal{V} = \{v_1, . . . , v_{|\mathcal{V}|}\}$ is the node set, $\mA$ represents adjacency matrix, and $\mX$ is the input node feature matrix. 
In TAGs, $\mX$ is usually generated from textual features via different models, e.g., language models. 
Additionally, $N(v)$ represents the neighboring nodes of node $v$.

\subsection{The \model Framework}

\vpara{Overview}
The overall framework of \model is illustrated in Figure ~\ref{fig:overview}. 
As mentioned previously, the focus of \model is on text-attributed graphs (TAGs), characterized by nodes described through text. 
In \model, we aim to harmonize the input features from disparate graphs into a unified distribution. 
This architecture consists of three coupled components for feature alignment across graphs. 
First, it employs a language model encoder to project the node texts from multiple graphs into a common semantic space. 
Subsequently, feature normalization is conducted on each graph independently, with the goal of consolidating graph clusters. 
Finally, we introduce an adaptive feature transformation mechanism, that is, mixture-of-feature-expert, which leverages dynamic routing to discern subtle nuances among the graphs.


\vpara{Language model as feature encoder}
Given the extensive applications of language models~\cite{achiam2023gpt,touvron2023llama} and various multimodal frameworks~\cite{achiam2023gpt,alayrac2022flamingo,li2023blip}, amount of features of different modalities, e.g., text, image, and numerical value, can be exactly expressed as textual descriptions, thereby enhancing the universality and adaptability of text-attributed graphs (TAGs). 
TAGs could be viewed as a bridge to connect graphs of different domains and facilitate the exploration of pretraining unified and transferable graph networks. 

Specifically, we utilize a language model (LM) as the encoder to convert all text features into continuous vectors, serving as the input features for all nodes on graphs. The vector representations for node $v_i$ are denoted as $\vx_i = LM(v_i)$. This LM-encoded input effectively encapsulates domain-specific information.
In this paper, we opt for E5-small~\cite{wang2022text}, noted for its exceptional proficiency in learning sentence embeddings, and for its simplicity and effectiveness. The embedding dimension is only 256 and  large-scale graphs with more than 100M nodes can be handled more efficiently. 

\begin{wrapfigure}{r}{0pt}
\centering
\begin{minipage}[t]{0.5\columnwidth}
\includegraphics[width=\linewidth]{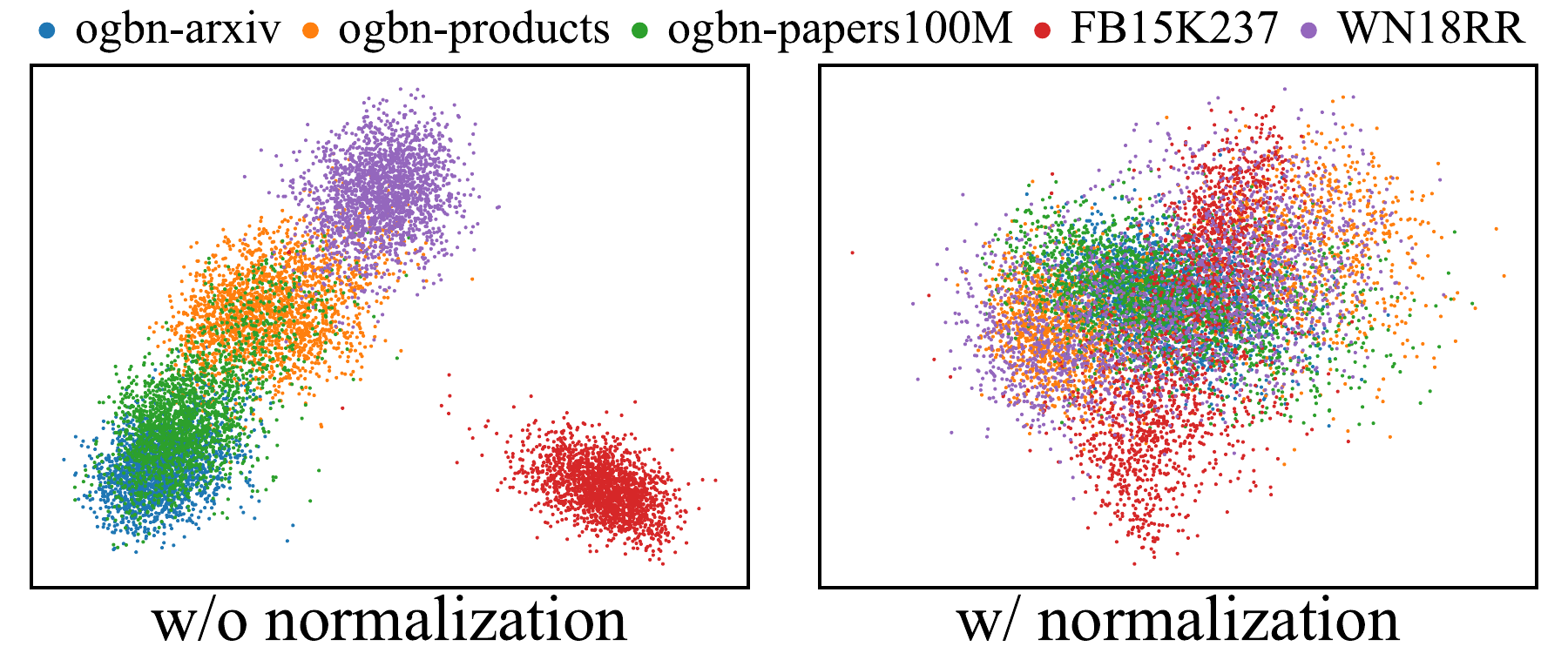}
\caption{Feature distribution comparison between \textit{w/} and \textit{w/o} normalization.}
\label{fig:feat_norm}
\end{minipage}
\end{wrapfigure}

\vpara{Feature normalization across graphs}
Now that language models map node features from different graphs into a shared semantic representation space, as shown in Figure~\ref{fig:feat_norm}, these features still fall into disparate clusters, which raises challenges due to the inherent diversity of these graphs in the following two aspects:

\begin{itemize}[leftmargin=*]
    \item \textit{Semantic dissimilarity within the same representation space.} For instance, the textual descriptions in academic and commercial networks vary significantly, leading to notable differences in their vector-space semantics. If the goal is to train a unified GNN that effectively processes tasks across graphs, the substantial input distribution disparities between these graphs can hinder the GNN's ability to discern commonalities. This discrepancy challenges the training of a GNN that performs as well or better than when individually trained on each graph due to the lack of mutual benefit across different graphs.
    \item \textit{Transferability of the pretrained GNN.} We aim to develop one GNN that is pre-trained to adapt to a wide range of tasks across different graphs, irrespective of their exposure during the training phase. To achieve this, it is imperative to closely align the feature distribution of out-of-distribution graphs with the training data's feature distribution.
\end{itemize}

To achieve these objectives, we attempt to reconcile and standardize the node feature distributions across different graphs. Inspired by the principle that latent embeddings in Variational Autoencoders (VAEs)~\cite{kingma2013auto} conform to a Gaussian distribution, we propose to normalize each graph individually to center their node feature distributions all around the zero point. Specifically, given the node features $\{\mX_1, ..., \mX_K\}$ of $K$ graphs, we perform the following normalization operation for each graph individually:
\begin{equation}
    \vx_j^{(k)} = \vx_j^{(k)} - \mu^{(k)}, \ ~ where \ \mu^{(k)} = \frac{1}{N} \sum \vx_j^{(k)}, ~ \vx_j^{(k)} \in \mX_{k} 
\end{equation}
where $\mu^{(k)}$ is the row-wise average of $\mX_{k}$. 
In this scenario, every graph feature displays an identical mean value close to zero. This uniformity facilitates the GNN pretraining in concentrating its learning efforts on the shared characteristics across graphs, rather than on modeling individual graph clusters independently. This approach might promote the discovery of transferable characteristics among the graphs. When using this pre-trained model on a previously unseen graph, the first step entails a normalization procedure that aligns the feature distribution accordingly to the pretraining setting.

In our approach, we only adjust the center of the node features without altering the standard deviation, a practice that diverges from the typical Gaussian normalization. This deviation arises because the language model encoder has already mapped the node embeddings to an appropriate distribution. Imposing a Gaussian distribution on these embeddings risks compromising semantic information, even within individual nodes in a graph.

\vpara{Mixture-of-feature-experts (MoF) with dynamic routing}
By the previously described techniques, node features across different graphs have been projected into a unified semantic space and fall into compact distributions around the zero point. Following this, it is imperative to further capture the nuance of feature distributions and further align features across multiple graphs.
A simplistic approach would involve using a distinct mapping function $F_i(\vx)$ to the node features of each graph. 
But this method would fail to generalize and transfer to graphs not seen during training and cannot establish inter-graph node relationships from feature normalization.

LM as the feature extractor and feature normalization presents two exploitable characteristics: 1) The feature dimensions of nodes across different graphs are consistent, and 2) The node embeddings from diverse graphs, distributed near the zero point, may exhibit overlapping regions. Furthermore, a pre-trained GNN is expected to capture inter-graph relationships and explore possibilities for mutual enhancement. This insight leads to the proposition of conducting node-level feature mapping, rather than graph-level, to capitalize on these characteristics.

Nevertheless, implementing node-level feature mapping is not a straightforward task. It necessitates a model that dynamically selects an appropriate feature mapper based on the input node features.
To achieve dynamic routing,
we propose a "Mixture-of-feature-experts" approach, analogous to the mixture-of-experts (MoE) design. This method leverages a combination of expert feature mappers, each tailored to handle specific aspects of node features across different graphs. By dynamically allocating nodes to the most suitable mapper based on their individual feature characteristics, this approach facilitates a more nuanced and effective integration of node features into a cohesive semantic space, thereby enhancing the pre-training of GNNs. 
Specifically, we predefine $m$ feature mappers ($m=4/8$ by default). Considering a node $v$ and its feature $\vx$, the process initiates with a gating layer. This layer is responsible for identifying the $topK$ experts that are most likely to be traversed by node $v$. 
$$ \mathrm{KeepTopK}(\vw, k) =\left\{
\begin{aligned}
w_i & \mathrm{~if} ~ w_i \mathrm{~ is ~ in ~ the ~ topk ~ elements} \\
-\infty & \mathrm{~otherwise}
\end{aligned}
\right.
$$

Then the $\vx$ is passed through the chosen $topK$ weighted experts to generate further-aligned features: 

\begin{equation}
    \begin{split}
        G(\vx) &= \mathrm{Softmax(KeepTopK}(\vx \cdot \mW_g, k)) \\
        \vh &= \sum_{i=1}^m G(\vx)_i F_i(\vx),
    \end{split}
\end{equation}
where $F_i(x) = \mW^0_i \vx, \mW^0_i\in \mathbb{R}^{d_0 \times d}$ and $\mW_g \in \mathbb{R}^{d\times K}$ are learnable weights. The outcome of this process is refined node features, which are subsequently fed into subsequent GNNs.


Though the proposed MoF resembles MoE (mixture-of-experts) to some extent, they exhibit two notable differences. Firstly, MoF employs the routing layer and mixture exclusively at the input layer, whereas MoE integrates them across all transformer layers. The primary aim of MoF is to amalgamate diverse features from various graphs, contrasting with MoE's objective of enlarging model capacity through sparse activation. Additionally, our experiments involving the extension of routing to multiple layers in MoF did not yield any incremental benefits. Second, MoF uses linear transformation as expert while MoE leverages MLP (Multi-layer perception) as the basic unit.

\subsection{Training and inference}
The overall training flow of \model is illustrated in Figure ~\ref{fig:overview}. During the preprocessing step, we first convert textual descriptions into continuous node features via a language model encoder and conduct feature normalization graph by graph. In the training stage, we first randomly sample batch graph nodes and derive their subgraphs via local clustering. The subgraphs first pass through a mixture-of-feature-experts and then a GNN encoder, e.g., GAT/GCN, follows to generate embeddings. After that, existing graph self-supervised methods
can all be applied for the pretraining.

\vpara{In-context inference}
In-context learning ~\cite{dong2022survey} aims to use an off-the-shelf pretrained model to solve novel tasks without the need for fine-tuning. And recently there are emerging works attempting to develop graph in-context learning~\cite{huang2023prodigy,liu2023one}. They either utilize an extra GNN to encode the context or design complex graph prompting strategies, which always bring extra cost.  

Our study, however, reveals that in-context learning can be achieved in a more straightforward manner using a pretrained GNN within a self-supervised framework. 
Specifically, in a few-shot learning scenario, given a query node $v_q$ and support set comprising $m$-class with $k$-samples in each class($m$-way,$k$-shot), their representations are denoted as $\vh_q$ and $\{\vh_{i,j}\}_{1 \leq i \leq m, 1 \leq j \leq k}$ respectively. 
To get the prediction of $v_q$, we simply average the representation of all nodes in each class, yielding the representative vector of this class: $\vy_i = \frac{1}{k}\sum_{j=1}^{k} \vh_{i,j}$. The classification of the query node $\vh_q$ is then determined by computing the cosine similarity between $\vh_q$ and $\{\vy_i\}_{1 \leq i \leq k}$ and identifying the class with the highest similarity as the predicted category for $v_q$:
\begin{equation}
    y_{q} = \mathrm{argmax}_{1\leq i \leq m} (\vh_q^\top \vy_i) / (\Vert \vh_q \Vert \cdot \Vert \vy_i \Vert ) 
\end{equation}

This simple design introduces almost no additional computational cost, trainable parameters, or prompt-engineering design but can achieve competitive performance under our pretraining setting. The results are shown in Table~\ref{tab:few_shot_citation}. 

\section{Experiment}
\label{section:Experiment}
In this section, we compare the effects of different graph SSL methods in joint and individual pretraining, demonstrating that the unified GNN obtained through \model can achieve performance improvements on various cross-domain datasets and is applicable to any graph SSL method.

\begin{table}[ht]
    \centering
    \footnotesize
    \caption{Linear probing results in unsupervised representation learning for node classification. \textmd{The model is pretrained on these datasets and we train a linear classifier to evaluate the embeddings generated from the pretrained GNN. We report Accuracy(\%). \textit{individually-pretrain} denotes that we train an individual GNN for each dataset. \textit{vanilla jointly-pretrain} represents training one GNN using all datasets without incorporating any designs.}
    }
    \renewcommand\arraystretch{1.05}
    \renewcommand\tabcolsep{6pt}
    \begin{tabular}{cccccc}
        \toprule[1.2pt]
         Method & Setting & ogbn-arxiv & ogbn-products & ogbn-papers100M & Avg. gain \\
         \midrule

        \multirow{1}{*}{MLP} 
         & supervised &  69.85$\pm$0.36  & 73.74$\pm$0.43 & 56.62$\pm$0.21 & -    \\
         
         \multirow{1}{*}{GAT}   
         & supervised &  74.15$\pm$0.15  & 83.42$\pm$0.35 & 66.63$\pm$0.23 & -   \\

         \multirow{1}{*}{GCN} 
         & supervised &    74.77$\pm$0.34    & 80.76$\pm$0.50 & 68.15$\pm$0.08  & -  \\

         \multirow{1}{*}{SGC} 
         & supervised & 71.56$\pm$0.41  & 74.36$\pm$0.27 & 58.82$\pm$0.08 & -   \\

         

        
        \midrule
         \multirow{3}{*}{BGRL} 
         &  individually-pretrain  & 72.98$\pm$0.14  & 80.45$\pm$0.16 & 65.40$\pm$0.23  & 0.0  \\
         & vanilla jointly-pretrain &69.00$\pm$0.08 &81.11$\pm$0.27 & 63.93$\pm$0.22 & -1.60 \\
         & \cellcolor{gray!20} \model & \cellcolor{gray!20} 73.20$\pm$0.20  & \cellcolor{gray!20} 80.79$\pm$0.45 & \cellcolor{gray!20} 65.62$\pm$0.14  & \cellcolor{gray!20} +0.26    \\
        \midrule


        \multirow{3}{*}{GRACE} 
         &  individually-pretrain & 73.33$\pm$0.19  & 81.91$\pm$0.27 & 65.59$\pm$0.13   & 0.0   \\
         & vanilla jointly-pretrain & 72.10$\pm$0.18 & 81.96$\pm$0.34 & 65.54$\pm$0.18 & -0.41 \\
         & \cellcolor{gray!20} \model & \cellcolor{gray!20} 73.69$\pm$0.26  & \cellcolor{gray!20} 81.90$\pm$0.19 & \cellcolor{gray!20} 65.61$\pm$0.17 & \cellcolor{gray!20} +0.12   \\

         \midrule

         \multirow{3}{*}{GraphMAE} 
         &  individually-pretrain & 72.35$\pm$0.12  & 81.69$\pm$0.11 & 65.68$\pm$0.28   & 0.0  \\
         & vanilla jointly-pretrain & 71.98$\pm$0.24 & 82.36$\pm$0.19 & 65.92$\pm$0.13  & +0.18 \\
         & \cellcolor{gray!20} \model & \cellcolor{gray!20} 72.97$\pm$0.22  & \cellcolor{gray!20} 82.51$\pm$0.18 & \cellcolor{gray!20} 66.08$\pm$0.18  & \cellcolor{gray!20} +0.61  \\

         \midrule
         \multirow{3}{*}{GraphMAE2} 
         &  individually-pretrain & 73.10$\pm$0.11  & 82.53$\pm$0.17 & 66.28$\pm$0.10   & 0.0  \\   
         & vanilla jointly-pretrain & 71.28$\pm$0.25  & 80.05$\pm$0.35 & 64.28$\pm$0.33 & -2.10 \\
         & \cellcolor{gray!20} \model  & \cellcolor{gray!20} 73.56$\pm$0.26  & \cellcolor{gray!20} 82.93$\pm$0.42 & \cellcolor{gray!20} 66.39$\pm$0.14 &  \cellcolor{gray!20} +0.32  \\ 

    \bottomrule[1.2pt]
    \end{tabular}
    \label{tab:main_exp}
\end{table}
\vspace{-5mm}

\subsection{Pre-training setting}

\vpara{Pre-training datasets.} 
The experiments are conducted on three public datasets of different scales, varying from hundreds of thousands of nodes to hundreds of millions. The statistics are listed in Table \ref{tab:statistics_dataset}. In the experiments, we follow the official splits in OGB~\cite{hu2020open} for ogbn-arxiv, ogbn-products, and ogbn-papers100M. To balance the data ratio and reduce training time, in ogbn-papers100M, we only use the nodes in the train/valid/test split and the nodes in the subgraphs corresponding to these nodes approximately 40 million in total for pre-training. 
The node features of the three datasets are generated by a language model (LM). Specifically, for ogbn-arxiv and ogbn-papers100M, we generate embeddings using the titles and abstracts corresponding to the nodes. For ogbn-products, we generate embeddings using the names and descriptions of the products associated with the nodes. The LM we use here is e5-small~\cite{wang2022text}.

\vpara{Pre-training methods.}
Our \model framework is flexible to the pre-training methods. 
In our experiment, we employ four types of self-supervised graph learning methods for mixed pre-training (\model): two contrastive methods, including GRACE~\cite{grace} and BGRL~\cite{BGRL}, as well as two generative methods, including GraphMAE~\cite{graphmae} and GraphMAE2~\cite{graphmae2}. 
These four methods represent the majority of self-supervised graph learning methods (SSL), demonstrating the versatility of \model with different SSL methods.

\subsection{Evaluation on linear probing}

To directly evaluate the performance of the pre-trained GNN model, we first use the linear probing method for evaluation, which is a widely used evaluation setting for self-supervised learning methods to judge the quality of the embeddings. 

\vpara{Setup.}
The datasets used are the same as the pre-trained ones, including ogbn-arxiv, ogbn-products, and ogbn-papers100M. The official splits of the three OGB node-level datasets are used for linear probing. The raw texts of the three datasets will be fed into the text encoder for node features. 
We report some supervised methods such as MLP, Graph Attention Network (GAT)~\cite{GAT}, Graph Convolution Network (GCN)~\cite{gcn}, and Simplified Graph Convolution (SGC)~\cite{SGC} to reflect the contributions of self-supervised learning.
For all baselines, we employ GAT~\cite{GAT} as the backbone for the encoder and the decoder. In the case of GRACE and BGRL, there is only the encoder.

\vpara{Evaluation.}
For linear probing, we first generate node embeddings with the pre-trained encoder. Then, we discard the encoder and train a linear classifier using the embeddings in a supervised setting. For the four SSL methods, each method undergoes three settings: \textit{``individually-pretrain''}, \textit{``vanilla jointly-pretrain''} and ``\model''. 
\textit{``individually-pretrain''} refers to pretraining individually on respective datasets and \textit{"vanilla jointly-pretrain"} means that three datasets are jointly trained in a straightforward way. 
We demonstrate the effectiveness of our method by comparing the results of \textit{``individually-pretrain''}, \textit{``vanilla jointly-pretrain''}, and \model across the four SSL methods mentioned above. We pre-train the GNN under three random seeds, with each seed running 10 trials of linear probing, and report the average accuracy and standard deviation. 

\vpara{Results.}
Table~\ref{tab:main_exp} summarizes the main results of the linear probing evaluation. 
Comparing individual pre-training and our mixed pre-training (\model), the latter performs better in most cases for different methods and datasets.
As for the vanilla jointly-pretrain method, it obtains worse performance than individual training in most cases.
Our \model achieves 1.86\%, 0.53\%, 0.43\%, and 2.42\% average improvements compared to the vanilla jointly-pretrain solution on three OGB datasets with BGRL, GRACE, GraphMAE, and GraphMAE2, respectively.  
From the perspective of self-supervised graph methods, the four methods perform differently, and GraphMAE2 achieves the best performance among the four methods. 
In general, under our \model framework, better graph SSL methods get better overall performance.

\setlength{\tabcolsep}{4mm}{
\begin{table}
    \centering
    \footnotesize
    \caption{Few-shot node classification results on ogbn-arxiv and Cora, and link classification results on WN18RR. \textmd{We report $m$-way-$k$-shot accuracy(\%), 5-way for ogbn-arxiv, Cora and WN18RR}
    } 
    \renewcommand\arraystretch{1.0}
    \renewcommand\tabcolsep{5pt}
    \begin{tabular}{lcccccc}
        \toprule[1.2pt]
        \multicolumn{1}{l}{}  & \multicolumn{2}{c}{ogbn-arxiv} & \multicolumn{2}{c}{Cora} & \multicolumn{2}{c}{WN18RR} \\
        \cmidrule{2-7}
          & 5-shot & 1-shot & 5-shot  & 1-shot & 5-shot & 1-shot  \\
         \midrule
         GPN  & 50.53{\tiny$\pm$3.07}  & 38.58{\tiny$\pm$1.61} & - & - & - & - \\
         TENT  & 60.83{\tiny$\pm$7.45}  & 45.62{\tiny$\pm$10.70} & - & - & - & - \\
         GLITTER & 56.00{\tiny$\pm$4.40}  & 47.12{\tiny$\pm$2.73} & - & - & - & - \\
         \midrule
         Prodigy & 61.09{\tiny$\pm$5.85} & 48.23{\tiny$\pm$6.18} & - & -  & - & - \\    
         OFA & 61.45{\tiny$\pm$2.56} & 50.20{\tiny$\pm$4.27}  & 48.76{\tiny$\pm$2.65}   & 34.04{\tiny$\pm$4.10}  & 46.32{\tiny$\pm$4.18} & 33.86{\tiny$\pm$3.41}   \\   
         \textit{OFA-emb-only} & 61.27{\tiny$\pm$7.09} & 43.22{\tiny$\pm$8.45}  & 58.60{\tiny$\pm$6.72}    & 40.87{\tiny$\pm$8.26}  &54.87{\tiny$\pm$9.73} &39.72{\tiny$\pm$9.35}   \\
         \midrule
         \model (GraphMAE) & 81.93{\tiny$\pm$6.22}  & 65.02{\tiny$\pm$10.62} & \textbf{74.49{\tiny$\pm$6.43}}   & \underline{55.55{\tiny$\pm$9.86}}  & \textbf{60.19{\tiny$\pm$10.31}} & \textbf{45.08{\tiny$\pm$10.55}}  \\       
         \model (GraphMAE2) & \underline{83.97{\tiny$\pm$5.85}} & \underline{70.65{\tiny$\pm$10.45}} & \underline{73.66{\tiny$\pm$6.75}}  &  \textbf{56.87{\tiny$\pm$9.98}} &  55.95{\tiny$\pm$10.49} & \underline{42.22{\tiny$\pm$10.04}} \\    
         \model (GRACE) & \textbf{84.76{\tiny$\pm$5.71}} & \textbf{71.18{\tiny$\pm$10.29}} & 69.85{\tiny$\pm$7.19}  &  52.60{\tiny$\pm$10.10} & 53.11{\tiny$\pm$10.24} & 39.58{\tiny$\pm$9.42}  \\ 
         \model (BGRL) & 81.88{\tiny$\pm$6.26}  & 66.31{\tiny$\pm$10.63} & 68.13{\tiny$\pm$6.84}   &  50.19{\tiny$\pm$9.49} & 51.97{\tiny$\pm$10.66} & 38.72{\tiny$\pm$9.77}   \\ 
         \textit{E5-emb-only} & 65.67{\tiny$\pm$7.02} & 47.13{\tiny$\pm$8.68}  & 59.71{\tiny$\pm$6.71}     & 41.58{\tiny$\pm$8.11} &\underline{56.52{\tiny$\pm$9.65}} &41.53{\tiny$\pm$9.36}     \\

    \bottomrule[1.2pt]
    \end{tabular}
    \label{tab:few_shot_citation}
\end{table}
}

\subsection{Evaluation on few-shot classification}
To better illustrate the transferable performance of the pre-trained model, we further conduct few-shot classification experiments for the model and evaluate the pretrained GNN on unseen graphs. 

\vpara{Setup.}
For the downstream few-shot classification tasks, we use Cora~\cite{yang2016revisiting} and ogbn-arxiv for the node-level task. 
Besides, we use two knowledge graphs for the link-level task, FB15K237~\cite{toutanova2015observed} and WN18RR~\cite{dettmers2018convolutional}, in the experiments to demonstrate the transferability of our pre-trained unified model. 
We compare our method with graph few-shot methods, including meta-learning methods, GPN~\cite{ding2020graph}, TENT~\cite{wang2022task}, GLITTER~\cite{wang2022graph}, Prodigy~\cite{huang2023prodigy}, and OFA~\cite{liu2023one}.

\vpara{Evaluation.}
Following the setting used in OFA~\cite{liu2023one}, we use 5-way-5-shot/1-shot for the few-shot node classification for evaluation. 
For few-shot link classification, we choose 5-way and 20-way evaluation settings for WN18RR and FB15K237, respectively. 
In-context learning is used for Prodigy, OFA, and our model.
Our \model and two embedding methods (i.e., OFA-emb-only and E5-emb-only) use a simple and effective solution for the few-shot evaluation. 
The averaged embeddings of nodes in the support set can be considered the corresponding classes' embeddings. 
The prediction of each query node will be the most similar class through the cosine similarity between the query embedding and the class embedding. 

\begin{table}[t]
    \centering
    \begin{minipage}[t]{0.47\textwidth}
        \centering
        \caption{Ablation of \model components on OGB datasets.\textmd{ \textit{- MoF} represents removing the MoF module and \textit{- Norm} denotes further excluding feature normalization.}}
        \scriptsize
        \renewcommand\tabcolsep{3pt}
        \begin{tabular}{clcccc}
            \toprule[1.2pt]
            Method & & Arxiv & Products & Papers100M & Avg. \\
            \midrule
            \multirow{3}{*}{GraphMAE}
            & \model & 73.17 & 82.45 & 66.23 &  \textbf{73.95} \\
            & - \textit{MoF} & 72.67 & 82.29 & 66.06 & 73.67 \\
            & - \textit{Norm} & 71.84 & 82.24 & 65.92 & 73.33 \\
            \midrule
            \multirow{3}{*}{GraphMAE2}
            & \model & 73.30 & 83.09 & 66.33 & \textbf{74.24} \\
            & - \textit{MoF} & 72.81 & 81.51 & 66.45 & 73.59 \\
            & - \textit{Norm} & 71.09 & 80.38 & 64.45 & 71.97 \\
            \midrule
            \multirow{3}{*}{GRACE}
            & \model & 73.74 & 82.06 & 65.67 & \textbf{73.82} \\
            & - \textit{MoF} & 73.38 & 81.80 & 65.43 & 73.54 \\
            & - \textit{Norm} & 72.08 & 82.23 & 65.61 & 73.31 \\
            \midrule
            \multirow{3}{*}{BGRL}
            & \model & 73.17 & 81.04 & 65.67 & \textbf{73.29} \\
            & - \textit{MoF} & 70.45 & 80.11 & 64.16 & 71.57 \\
            & - \textit{Norm} & 69.05 & 81.39 & 63.71 & 71.38 \\
            \bottomrule[1.2pt]
        \end{tabular}
        \label{tab:ablation_withoutmoe_notrick}
    \end{minipage}%
    \hspace{0.04\textwidth}
    \begin{minipage}[t]{0.47\textwidth}
        \centering
        \caption{Ablation on the transferability across different OGB datasets. \textmd{We pretrain the GNN on only-one graph and test the performance on all datasets. "ogbn-" is omitted for brevity.}}
        \scriptsize
        \renewcommand\tabcolsep{1pt}
        \renewcommand{\arraystretch}{1.2}
        \begin{tabular}{clcccc}
            \toprule[1.2pt]
            & & \multicolumn{4}{c}{Pretrain on} \\
            Method & & \makecell{Arxiv\\only} & \makecell{Products\\only} & \makecell{Papers100M\\only} & \model \\
            \midrule
            \multirow{4}{*}{\makecell{\tiny GraphMAE \\ Eval on}}
            & Arxiv & 72.42- & 72.68\textcolor[RGB]{199, 54, 44}{$\uparrow$}& 73.03\textcolor[RGB]{199, 54, 44}{$\uparrow$} & 73.17\textcolor[RGB]{199, 54, 44}{$\uparrow$} \\ 
            & Products & 78.11\textcolor[RGB]{44, 138, 60}{$\downarrow$}& 81.64- & 81.93\textcolor[RGB]{199, 54, 44}{$\uparrow$} & 82.45\textcolor[RGB]{199, 54, 44}{$\uparrow$} \\
            & Papers100M & 63.56\textcolor[RGB]{44, 138, 60}{$\downarrow$}& 65.93\textcolor[RGB]{44, 138, 60}{$\downarrow$}& 66.03- & 66.23\textcolor[RGB]{199, 54, 44}{$\uparrow$} \\
            \cmidrule{2-6}
            & {\textit Average} & 71.36 & 73.42 & 73.66 & \textbf{73.95} \\
            \midrule
            \multirow{4}{*}{\makecell{\tiny GraphMAE2 \\ Eval on}}
            & Arxiv & 73.00- & 72.76\textcolor[RGB]{44, 138, 60}{$\downarrow$}& 72.09\textcolor[RGB]{44, 138, 60}{$\downarrow$}& 73.30\textcolor[RGB]{199, 54, 44}{$\uparrow$} \\
            & Products & 80.23\textcolor[RGB]{44, 138, 60}{$\downarrow$}& 82.43- & 81.53\textcolor[RGB]{44, 138, 60}{$\downarrow$}& 83.09\textcolor[RGB]{199, 54, 44}{$\uparrow$} \\
            & Papers100M & 64.59\textcolor[RGB]{44, 138, 60}{$\downarrow$}& 65.47\textcolor[RGB]{44, 138, 60}{$\downarrow$}& 66.30- & 66.33\textcolor[RGB]{199, 54, 44}{$\uparrow$} \\
            \cmidrule{2-6}
            & {\textit Average} & 72.61 & 73.55 & 73.31 & \textbf{74.24} \\
            \bottomrule[1.2pt]
        \end{tabular}
        \label{tab:ablation_cross_test_mae2}
    \end{minipage}
\end{table}

\vpara{Results.}
Table~\ref{tab:few_shot_citation} shows the main few-shot results. And the result of FB15K237 is shown in Appendix \ref{appendix:FB15K237 few-shot classification result}.
For the node-level task, our framework with each self-supervised method significantly outperforms existing few-shot methods.
We also validate the advantage of the in-context inference used in our framework. The results of OFA-emb are obtained using the node embeddings of OFA and our in-context inference strategy. On the Cora dataset, the OFA-emb performs even better than the original OFA. 
We also evaluate the embedding of the E5 model, which is used in our framework to model the raw texts. The E5-emb performs better than the OFA-emb in all cases except FB15K237. 
As for link classification on knowledge graphs, our method achieves better results than OFA on WN18RR with 5-/1-shot scenarios. 
Note that our model does not see any information about knowledge graphs in the pre-training, which shows the strong transferability of our model.

\begin{figure}[H]
    \centering
    \includegraphics[width=0.3\textwidth]{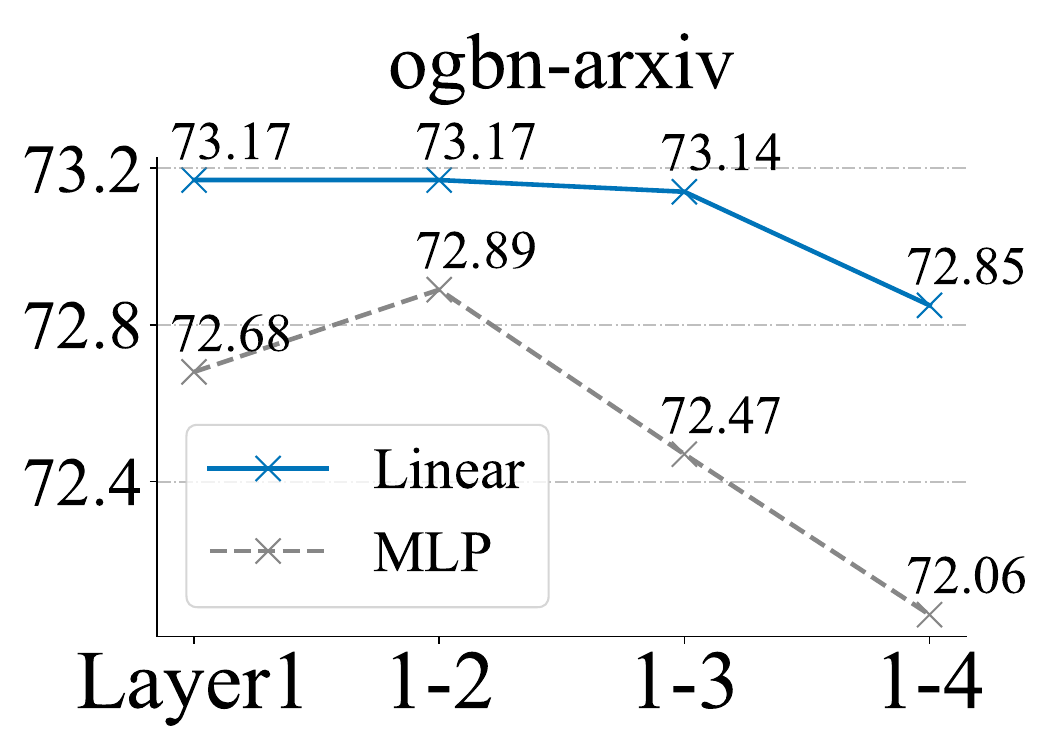}
    \includegraphics[width=0.3\textwidth]{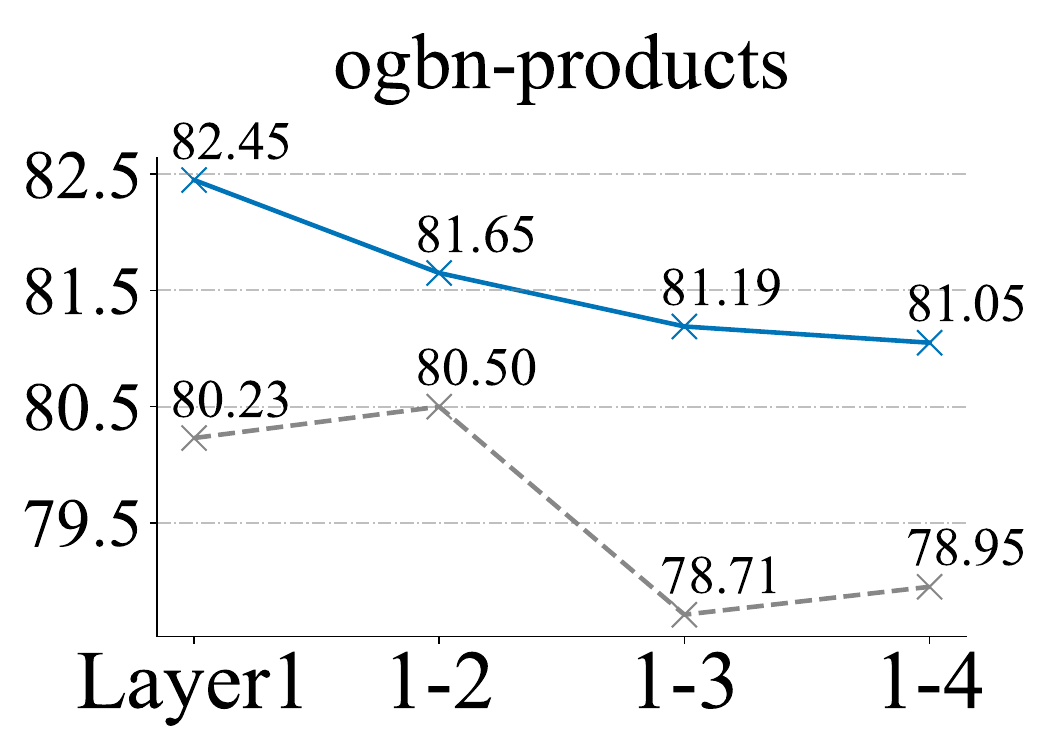}
    \includegraphics[width=0.3\textwidth]{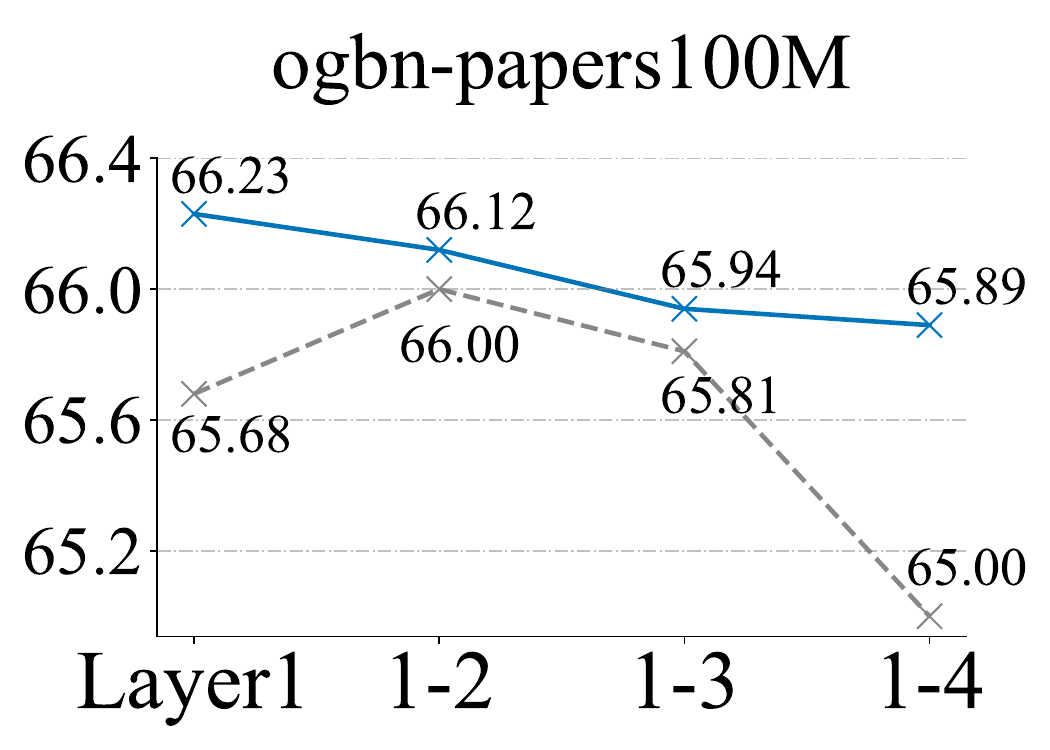}
    \caption{Ablation on the number of MoF layers and MoF projectors. \textmd{ \textit{Linear} represents using linear transformation as projectors and \textit{MLP} represents using multi-layer perception as projectors. Using \textit{Linear} and employing MoF at the input layer is the best.}}
    \label{fig:ablation_MOE_layer_unit}
\end{figure}

\subsection{Ablation Studies}
\vpara{Ablation on \model components.}
We investigate the effects of different components in \model design, i.e., feature transformation and MoF module. The results are illustrated in Table~\ref{tab:ablation_withoutmoe_notrick}. 
Excluding the MoF module and feature normalization both harm the performance across various datasets and graph SSL methods. Notably, GraphMAE2 and BGRL show greater sensitivity whereas GraphMAE and GRACE are more stable in joint-training settings. This suggests that the proposed feature alignment can effectively enable GNNs to capture commonalities across different graphs, promoting mutual benefits.

\vpara{Ablation on MoF design.} 
We study the influence of MoF design.
The results of GraphMAE are shown in Figure ~\ref{fig:ablation_MOE_layer_unit}, 
illustrating a progressive increase in the number of GNN layers using MoF from the input layer to all layers. 
It is observed that using linear transformation (Linear) shows a better performance than multi-layer perception (MLP) as feature transformation. And attaching more MoF layers to the GNN generally brings a performance drop in all datasets.

\vpara{Ablation on transferability across datasets.} 
We pretrain a GNN on one dataset and evaluate the GNN on all datasets to test the transferability among different OGB datasets~\ref{tab:main_exp}. 
Moreover, within the GraphMAE framework, pre-training on the ogbn-papers100M dataset consistently yields benefits in downstream tasks. However, GraphMAE2 exhibits limited transferability across all datasets, suggesting that GraphMAE may more effectively integrate or learn shared characteristics. 
And \model can benefit both methods across all datasets, showing its advantage.

\section{Conclusion}

In this work, we aim to develop a framework to pretrain one GNN model that can be applied across diverse graph domains. To achieve this, we propose \model to integrate different graphs via feature alignment. By incorporating a language model as the feature encoder, we further devise feature normalization and a mixture-of-feature-expert module to align feature distributions. Extensive experiments show that the proposed \model can seamlessly integrate with existing graph SSL methods and show promising performances on linear probing and few-shot classification tasks on in-domain and out-of-domain data.

\vpara{Limitations} 
\label{section: Limitations}
Despite extensive experiments and promising justifications, our method has several limitations:
1) The experiments are primarily conducted on textual graphs, as commonly used graph datasets predominantly contain text features. It would be beneficial to collect graphs from more modalities to further verify the effectiveness of \model. 2) In the future, we aim to explore more theoretical insights into the feature distribution changes.

\bibliographystyle{plain}
\bibliography{reference}


\clearpage
\appendix

\section{Dataset Statistics}
We provide the details of dataset statistics used in our experiments in Table\ref{tab:statistics_dataset}.

\begin{table}[ht]
    \centering
    \caption{Statistics of datasets.}
    \begin{tabular}{l|ccrr}
    \toprule[1.2pt]
        Datasets &Domain &Task & \#Nodes & \#Edges  \\
    \midrule
        ogbn-arxiv &Citation & Node & 169,343 & 1,166,243 \\
        ogbn-papers100M &Citation & Node & 111,059,956 & 1,615,685,872 \\
        ogbn-products &Product & Node & 2,449,029 & 61,859,140 \\
    \midrule
        Cora & Citation & Node & 2,708 & 10,556 \\
        FB15K237 & Knowledge & Link &14,541 & 310,116  \\
        WN18RR & Knowledge & Link & 40,943 & 93,003 \\
    \bottomrule[1.2pt]
    \end{tabular}
    
    \label{tab:statistics_dataset}
\end{table}

\section{Additional Experimental Results}
\subsection{FB15K237 few-shot classification result}
\label{appendix:FB15K237 few-shot classification result}
Table~\ref{tab:few_shot_FB15K} shows the few-shot classification results of FB15K237. We report 20-way 5-shot and 1-shot accuracy(\%) for FB15K237. 
In the Prodigy setting, GNN is pretrained on a large-scale knowledge graph(Wiki) constructed from Wikipedia, and in the One-For-All setting, GNN is directly trained on FB15K237. However, in our setting, our model does not see any information about knowledge graphs. In this case, our model still outperforms Prodigy by 5\% in 5-shot and 8\% in 1-shot, which shows the strong transferability of our model.

\begin{table}[ht]
    \centering
    \caption{Few-shot link classification results on FB15K237. }
    \begin{tabular}{lcc}
        \toprule[1.2pt]
        \multicolumn{1}{l}{}  & \multicolumn{2}{c}{FB15K237} \\
        \cmidrule{2-3}
         & 5-shot  & 1-shot \\
       
         \midrule
         Prodigy  & 74.92{\tiny$\pm$6.03} & 55.49{\tiny$\pm$6.88} \\    
         OFA   & 82.56{\tiny$\pm$1.58} & 75.39{\tiny$\pm$2.86}   \\   
         \textit{OFA-emb-only}   & 59.11{\tiny$\pm$6.95}   & 43.03{\tiny$\pm$7.17}  \\
         \midrule
         \model (GraphMAE)  & 79.92{\tiny$\pm$5.54} & 63.01{\tiny$\pm$7.29}  \\       
         \model (GraphMAE2)  & 79.86{\tiny$\pm$5.53}  & 63.56{\tiny$\pm$7.31} \\    
         \model (GRACE)  & 75.04{\tiny$\pm$5.98}  & 60.09{\tiny$\pm$7.36} \\ 
         \model (BGRL)  & 77.74{\tiny$\pm$5.87}  & 61.48{\tiny$\pm$7.44}  \\ 
         \textit{E5-emb-only} & 58.43{\tiny$\pm$6.94}   & 42.06{\tiny$\pm$7.11}   \\

    \bottomrule[1.2pt]
    \end{tabular}
    \label{tab:few_shot_FB15K}
\end{table}

\subsection{Ablation on hyper-parameters sensitivity}
We study the hyper-parameter's influence on our method. The result in Table\ref{tab:high-parameters sensitivity} illustrates that the performance of GraphAlign is relatively stable and less sensitive to hyper-parameters. Specifically, We conducted experiments on 6 hyper-parameters: learning rate, epochs, number of experts, number of top k, number of GNN layers, and number of hidden sizes.

\begin{table}
\centering
\small
\caption{Ablation on high-parameters sensitivity. \textmd{The graph SSL method is GraphMAE and the hyperparameters of the GraphAlign reported in the Table\ref{tab:main_exp} are lr 0.0002, epoch 20, number of experts 4, topk 1, GNN layers 4, and hidden sizes 1024.}}
\renewcommand\tabcolsep{7pt}
\begin{tabular}{lcccccccccc}
\toprule
 & \multicolumn{4}{c}{Lr} & \multicolumn{5}{c}{Epoch} \\
\cmidrule(lr){2-5} \cmidrule(lr){6-10}
Dataset & 1e-4 & 2e-4 & 5e-4 & 1e-3 & 5 & 10 & 15 & 20 & 25 \\
\midrule
ogbn-arxiv & 72.66 & 73.17 & 72.69 & 72.72 & 72.91 & 72.67 & 73.04 & 73.17 & 73.15 \\
ogbn-products & 82.31 & 82.45 & 81.49 & 81.39 & 81.15 & 82.21 & 82.32 & 82.45 & 82.44 \\
ogbn-papers100M & 66.26 & 66.23 & 65.97 & 66.08 & 65.97 & 65.86 & 66.13 & 66.23 & 66.22 \\
\midrule
Avg. & 73.74 & \textbf{73.95} & 73.38 & 73.40 & 73.34 & 73.58 & 73.83 & \textbf{73.95} & 73.94 \\
\bottomrule
\end{tabular}

\bigskip 
\renewcommand\tabcolsep{6pt}
\begin{tabular}{lcccccccccc}
\toprule
 & \multicolumn{3}{c}{Num of Expert} & \multicolumn{3}{c}{Num of k} &\multicolumn{2}{c}{Layers} &\multicolumn{2}{c}{Hidden size} \\
\cmidrule(lr){2-4} \cmidrule(lr){5-7} \cmidrule(lr){8-9} \cmidrule(lr){10-11}
Dataset & 3 & 4 & 5 & 1 & 2 & 4 & 2 & 4 & 512 & 1024 \\
\midrule
ogbn-arxiv & 72.93 & 73.17 & 72.73 & 73.17 & 72.95 & 72.81 & 73.30 & 73.17 & 72.16 & 73.17 \\
ogbn-products & 81.86 & 82.45 & 82.15 & 82.45 & 81.87 & 81.50 & 80.84 & 82.45 & 82.13 & 82.45 \\
ogbn-papers100M & 65.97 & 66.23 & 66.12 & 66.23 & 66.16 & 66.09 & 65.48 & 66.23 & 65.76 & 66.23 \\
\midrule
Avg. & 73.59 & \textbf{73.95} & 73.67 & \textbf{73.95} & 73.66 & 73.47 & 73.21 & \textbf{73.95} & 73.35 & \textbf{73.95} \\
\bottomrule
\label{tab:high-parameters sensitivity}
\end{tabular}
\end{table}

\subsection{Ablation on GNN backbone model}
To prove our GraphAlign works on different GNNs. We use GCN as another GNN backbone model in GraphMAE. The result in Table\ref{tab:GNN backbone model} illustrates that GraphAlign applies to different GNN backbone models.

\begin{table}
\centering
\caption{Ablation on GNN backbone model.}
\renewcommand{\arraystretch}{1.2}
\begin{tabular}{lcc}
\toprule[1pt]
 & \multicolumn{2}{c}{GraphMAE} \\
\cmidrule(lr){2-3} 
Dataset &   Individually-Pretrain &  GraphAlign    \\
\midrule
ogbn-arxiv & 73.34 & 74.39  \\
ogbn-products & 80.25 & 80.77  \\
ogbn-papers100M & 65.54 & 65.89  \\
\bottomrule[1pt]
\end{tabular}
\label{tab:GNN backbone model}
\end{table}

\hide{
\begin{table}[ht]
\centering
\caption{Ablation on GNN backbone model.}
\renewcommand\tabcolsep{2pt}
\small
\begin{tabular}{lcccccc}
\hline
 & \multicolumn{2}{c}{Graphmae} & \multicolumn{2}{c}{Graphmae2} & \multicolumn{2}{c}{Grace} \\
\cmidrule(lr){2-3} \cmidrule(lr){4-5} \cmidrule(lr){6-7} 
Dataset & {\scriptsize  Individually-Pretrain} &  {\scriptsize Graphalign} & {\scriptsize  Individually-Pretrain} &  {\scriptsize Graphalign} & {\scriptsize  Individually-Pretrain} &  {\scriptsize Graphalign}   \\
\hline
ogbn-arxiv & 73.34 & 74.39 & xx & xxx & xx & xx \\
ogbn-products & 80.25 & 80.77 & xx & xxx & xx & xx \\
ogbn-papers100M & 65.54 & 65.89 & xx & xxx & xx & xx \\
\hline
\end{tabular}
\label{tab:GNN backbone model}
\end{table}
}

\section{Experimental Details}
\subsection{Implementation details}
\label{section: Computer resources}
The experiments are conducted on a Linux machine with 1007GB RAM, and 8 NVIDIA A100 with 40GB
GPU memory. As for software versions, we use Python 3.9, PyTorch
1.12.0, OGB 1.3.3, and CUDA 11.3. 
The whole experiment can be done on one single A100. For instance, using GraphMAE to pretrain GNN on three OGB datasets, it will need 70GB RAM, 26GB GPU memory and 26 hours (batch size 512 and epoch 20). Our code supports distributed training, you can use two A100 to train the GNN within 13 hours, 130GB RAM.

\subsection{Complexity analysis}
Below is the computational and space complexity of the three components of GraphAlign(feature generation, normalization, and MoF). 
Feature generation and normalization are conducted as a preprocessing step and are once-for-all work. For feature generation, we simply pass all nodes through a language model to get the embedding. So the complexity of feature generation and normalization are both $O(N)$, where $N$ is the number of nodes and the space complexity is also $O(N)$.
As for MoF, the added MoF layer only involves dense matrix multiplication, and thus the complexity is $O(Bd^2)$, where $B$ is the batch size and $d$ is the hidden dimension. In comparison, the computational complexity of GNN backbone (GCN for example) could be simply denoted as $O(LEd + LBd^2)$, where $L$ is the number of layers and $E$ is the number of edges in the sampled subgraph. So MoF brings little additional computational cost.

\subsection{Subgraph sampling for large graph training}

\vpara{Instance selection and sampling}
Our aim is to pretrain on multiple large-scale graphs, and sampling is necessary as it is infeasible to load all graphs into GPUs due to memory limits. To facilitate the pretraining of the aforementioned procedure on multiple graphs, we view the subgraph as the fundamental unit, akin to images in computer vision and sentences in natural language processing. This also ensures compatibility with all existing graph self-supervised learning algorithms, both generative~\cite{graphmae,graphmae2} and contrastive~\cite{BGRL,grace,zheng2022rethinking}.

Node-wise sampling strategy, which involves sampling a subgraph given each query node, is an ideal option as it allows for the flexible adjustment of the influence from various datasets by controlling the number of subgraphs sampled from each dataset. 
In this work, we use local clusters~\cite{andersen2006local,spielman2013local} to obtain subgraphs. We run the PPR (Personalized PageRank) algorithm to derive nodes with the top-$M$ highest PPR scores ($M=256/512$ by default) and sample a subgraph consisting of these nodes for SSL training. This procedure can be efficiently implemented utilizing Approximate-PPR as detailed in ~\cite{andersen2006local}. The complexity of this method is $\mathcal{O}(\frac{1}{\epsilon})$, with $\epsilon$ representing a small constant.

\end{document}